\documentclass[11pt,a4paper,twocolumn]{article}
\usepackage[margin=1in]{geometry}
\usepackage{authblk}
\usepackage{blindtext}
\usepackage{amsfonts}
\usepackage[center]{caption}
\usepackage{abstract}
\usepackage[mathscr]{eucal}
\usepackage{booktabs}
\usepackage{graphicx}
\usepackage{amsmath}
\usepackage{float}
\usepackage{url}
\newcounter{biburlucpenalty}
\newcounter{biburllcpenalty}
\usepackage{breqn}
\usepackage{microtype}
\DeclareMathOperator*{\argmax}{arg\,max}
\usepackage{etoolbox}
\patchcmd{\thebibliography}{\section*{\refname}}{}{}{}
\title{Maximum Entropy Dueling Network Architecture in Atari Domain}
\author[1]{Alireza Nadali}
\author[1]{Mohammad Mehdi Ebadzadeh}

\affil[1]{Department of Computer Engineering, Amirkabir University of Technology, No. 350, Hafez Ave, Valiasr Square, Tehran, Iran}
\affil[]{\textit{ \{A\_Nadali, Ebadzadeh\}@aut.ac.ir }}

\date{}
\begin{document}
\maketitle
\vspace{\fill}
\begin{abstract}
In recent years, there have been many deep structures for Reinforcement Learning, mainly for value function estimation and representations. These methods achieved great success in Atari 2600 domain. In this paper, we propose an improved architecture based upon Dueling Networks, in this architecture, there are two separate estimators, one approximate the state value function and the other, state advantage function. This improvement based on Maximum Entropy, shows better policy evaluation compared to the original network and other value-based architectures in Atari domain.
\end{abstract}

\section{Introduction}

In the course of recent years, Deep Learning made great contributions in the field of Artificial Intelligence and Machine Learning \cite{Lecun2015}. In Reinforcement Learning, momentous examples are Deep Q-Learning \cite{Mnih2015}, Deep Visuomotor Policies \cite{Levine2016}, massively parallel frameworks \cite{Nair2015} and AlphaGo \cite{Silver2016}, which beaten humanity at our own games. These algorithms employ standard neural networks, such as MLPs and Convolutional networks, in order to achieve their results. The main focus of researchers in recent years have been to improve these standard networks and tackle great many difficulties posed by RL methods. Dueling Network \cite{Wang2016} took a rather different approach, Wang et al decided to alter the architecture in an innovative way in order to outperform the state-of-the-art methods.

In this paper, we propose new method based on Dueling Networks, which tries to improve mentioned DRL architecture. The rest of the paper is organized as follows. Section 2 briefly reviews the literature, Section 3 describes preliminaries for this research, Section 4 portraits the theoretical aspect of our method, Section 5 is the results and finally the paper is concluded in Section 6.
\section{Literature Review}
The primary goal of researchers in the field of Artificial Intelligence is to design fully autonomous agents that can interact with the environment and learn optimal behaviors, improving ever so slightly just like humans. Designing such an agent is a daunting task, though researchers in the past have had success creating such agents \cite{Silver2016,Singh2002}, these agents are limited to fairly low-dimensional domains. What plagues the traditional RL approaches is memory, time and sample complexity \cite{Strehl2006}. In recent years however, rise of Deep Learning made it possible to employ powerful function approximators to overcome these drawbacks. 

Deep Learning dramatically changed many areas in Machine Learning such as object detection, speech recognition and computer vision \cite{Lecun2015}. The most important aspect of Deep Learning that allows us to perform tasks that were previously impossible, is the ability to automatically represent high-dimensional data (e.g., images, text and audio) in low-dimensional domain. This property is achieved using deep neural networks and architecture that almost eliminates the curse of dimensionality \cite{Bengio2013}. RL is no exception, with the Deep Learning as a powerful tool, a new term has been coined by researchers which is "Deep Reinforcement Learning". Deep Learning enables RL to solve real-world problems that otherwise would be intractable. Amongst the recent innovations in DRL, the first that revolutionized and changed the course of RL related researches was development of an algorithm that can learn to play wide range of Atari 2600 games on superhuman level, just by extracting features from images \cite{Mnih2015}. This research was the first to demonstrate the potency of RL algorithms combined with Deep Learning, which can train agents only with raw and relatively high-dimensional images. AlphaGo is another agent which managed to beat human world champion in Go, it is a hybrid system that combines advanced search trees and deep neural network \cite{Silver2016}. Neural networks in AlphaGo were trained using supervised and reinforcement learning, in tandem with a traditional heuristic search algorithm.

DRL algorithms have already achieved great success in wide range of problems such as robotics, where control policy for robots is learned directly from input images of the real world \cite{Levine2016,Levine2017}, which outperform controllers that were previously designed using Control Theory. These algorithms are so capable that researchers have been able to design agents with meta-learning ability (learn to learn) \cite{He2017,To2019}.

Domain of video games is indeed intriguing, but it's not the end goal for DRL algorithms. Main focus behind DRL is to design agents that can learn and adapt in real world situations \cite{Luo2014,Levine2017}, having said that, RL is also an interesting way of approaching optimization problems using trial and error \cite{Zoph2017,Li2017}. DRL methods have also been applied to traditional RL methods, scaling up prior work to high-dimensional problems, as we previously mentioned, DRL can address curse of dimensionality spectacularly \cite{Bengio2013}. the well known function approximation ability of deep architectures made them a great choice for regressing various RL functions (that we discuss in the next section), one of the earliest implementation is TD-Gammon, neural network that reached champion-level performance in Backgammon decades ago \cite{Tesau1995}. Current methods, address the highly complex domain of inputs such as images and videos \cite{Mnih2015,Schulman2015,Mnih2016,Lange2012,Fang2021}.  One of the key components of RL is Q-function, van Hasselt showed that the single estimator in Q-learning is suffering from over-estimation, hence a new algorithm with double estimator was proposed that improved the learning process immensely \cite{VanHasselt2010}. Though Double Q learning requires one additional function to be learned, but researchers proposed a way to use available target  network from DQN to reduce the complexity and achieve better results \cite{Xiao2020}.
\section{Preliminaries}
In this section, we will introduce the field of RL and mathematical formulations. Essentially, RL agent learns through interaction, it interacts with the environment, observes the reward for its actions and can learn to optimize its behaviour by the reward it received. This method of learning has its roots in psychology, which is the foundation of RL \cite{Morales2011}. 

\begin{figure*}[!ht]
\centering
\hbox{\hspace{1.5em} \includegraphics[scale=0.8]{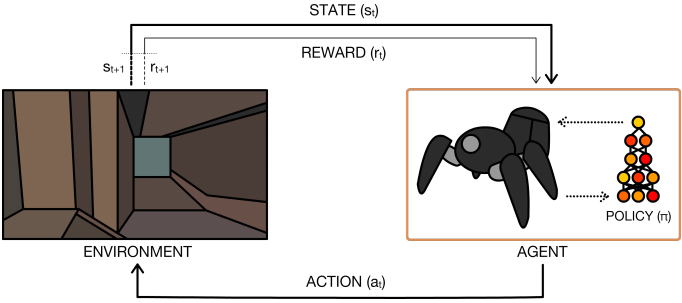}}
\caption{\footnotesize {\textit{The perception-action-learning-loop}, at the timestep \textit{t}, agent receives the reward r\textsubscript{t} and observes state s\textsubscript{t}, the agent uses its policy to determine which action a\textsubscript{t} yields the best rewards, then executes this action and environment transitions to the new state s\textsubscript{t+1}. Agent uses the state transition tuple in the form of (s\textsubscript{t}, a\textsubscript{t}, s\textsubscript{t+1}, r\textsubscript{t+1}) to improve its policy \cite{Arulkumaran2017ABS}.
} }

\label{fig1}
\end{figure*}

Formally, an \textit{agent}, following an algorithm, observes a \textit{state} s\textsubscript{t} from the \textit{environment}, at timestep t. Then, agent interacts with the environment by performing an action a\textsubscript{t}, the environment transitions to the next state s\textsubscript{t+1} which is dependent on current state s\textsubscript{t} and action a\textsubscript{t}. Agent receives a \textit{reward} r\textsubscript{t+1} from the environment as feedback. The goal of RL is to learn a \textit{policy} $\pi$ that maximizes the expected return of the reward function. Given a state, policy returns an action for agent to take, an \textit{optimal policy} maximizes the expected cumulative reward of the sequence. Every action taken by the agent contains useful information that  agent utilizes to learn, which is illustrated in Figure \ref{fig1}.

\subsection{\small{Markov Decision Processes}}
RL can be modeled as a Markov Decision Process (MDP), under these assumptions:
\begin{itemize}
\item{A set of states $\mathcal{S}$ and probability of the starting state $p(s_0)$.}
\item{A set of actions $\mathcal{A}$.}
\item{$\mathcal{T}$(s\textsubscript{t+1}$|$ s\textsubscript{t}, a\textsubscript{t})} Transition dynamics, which is the mathematical model of the system that maps current state and action to the next state.
\item{Reward function r\textsubscript{t}.}
\item{discount factor $\gamma\in [0,1$] in order to put emphasis on immediate rewards and make sum of the rewards tractable.}
\end{itemize}
Generally, the policy $\pi$ maps states to a new distribution over actions: $\pi : \mathcal{S} \to p(\mathcal{A}=a|s)$. Every sequence of states, actions and rewards is an\textit{trajectory} of the given policy, which returns a reward function $R=\sum_{t=0}^{T-1}\gamma^{t}r_{t+1}$(T is the last time step). RL aims to find an optimal policy that maximizes the reward function:
\begin{equation}
\pi^*=\argmax_\pi \mathbb{E}[R|\pi]
\end{equation}
A key assumption in MDPs is that the current state is only dependent on the previous state and action, under the D-separation property, rather than all previous states. Having said that, this assumption is unrealistic as it requires the states to be \textit{fully observable}, which does not hold true for every problem. 
\subsection{\small{Value and Quality Functions}}
The state-action quality function $Q^{\pi}$(s,a) is defined as expected value of the reward function given the state action tuple under a certain policy:
\begin{equation}
Q^\pi(s,a) = \mathbb{E}[R|s,a,\pi]
\end{equation} 
The state value function $V^{\pi}(s)$ is the expected reward function given state under a certain policy:
\begin{equation}
V^\pi(s) = \mathbb{E}[R|s,\pi]
\end{equation}
it can be rewritten as:
\begin{equation}
V^\pi(s_t) = \mathbb{E}_{a_t\sim \pi(a_t,s_t)}[Q^{\pi}(s_t,a_t)]
\end{equation}
One may say Value function determines how good or bad a certain state is, by summing over all possible actions, hence, optimal policy's value function is:
\begin{equation}
V^*(s)=\max_\pi V^\pi(s)\qquad \forall s \in \mathcal{S}
\end{equation}
In order to learn the Q function, we employ Markov property and rewrite the function using a Bellman equation \cite{Bajic2003}, which estimates Q function recursively:
\begin{equation}
Q^\pi(s_t,a_t)=\mathbb{E}_{s_{t+1}}[r_{t+1}+\gamma Q^\pi(s_{t+1},\pi(s_{t+1})]
\end{equation}
\begin{figure*}[hbt!]
\centering
\hbox{\hspace{1.5em} \includegraphics[scale=0.6]{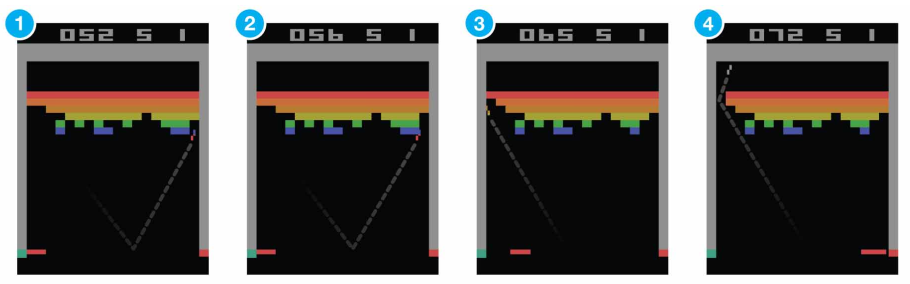}}
\caption{\footnotesize {Four frames of Atari 2600 game \textit{Breakout} \cite{Mnih2015}.}}
\label{fig2}
\end{figure*}
Meaning $Q^\pi$ can be improved upon by a method called \textit{bootstrapping}, i.e., one can employ current estimated values of $Q^\pi$ to further improve the Q function. This is the bedrock of Q-learning \cite{Watkins1992} and SARSA algorithm \cite{Rummery1994}.

\subsection{\small{Atari 2600 Domain}}
So far, we've briefly reviewed notable literature and presented the mathematical foundation of RL, here we describe the Atari 2600 games and how the data set is formed. 

This data set consist of raw Atari 2600 frames, which are $210 \times 160$ pixel images with a 128-color palette. Firstly, a preprocessing step is applied aiming at dimensionality reduction and to deal with artefacts of Atari 2600 emulator (which is caused by limitations of Atari 2600 sprites). First step is to take the maximum value for each pixel color over the current and previous frame. second step is to extract the Luminance channel (more commonly known as Y channel) from RGB and rescale it to $84 \times 84$. Lastly, we stack the 4 most recent frames in order to model time dependency. Figure \ref{fig2} shows four different frames of \textit{Breakout} which was a popular Atari game. 

\subsection{\small{Deep Q-networks}}
Here, we go in detail of how neural networks are employed to approximate Q and value functions. First of all, we need to define another important function which is the key to understanding Dueling networks. \textit{Advantage function} is calculated as follows:
\begin{equation}
A^\pi(s,a) = Q^\pi(s,a) - V^\pi(s)
\end{equation}
which means $\mathbb{E}_{a\sim \pi(s)}[A^\pi(s,a)]=0$. Value function is a measure of how good a state is, Q function on the other hand, tells us the relative importance of each action. Advantage function tells us how beneficial one action is with respect to the state s. 

Value function represented in previous sections is typically high-dimensional and is the ideal function to approximate using neural networks. Let us define $Q(s,a;\theta)$ as the estimated value by neural networks with parameters $\theta$. In order to train and optimize this network, we define the loss function at $i$\textsuperscript{th} step as follows:
\begin{equation}
L_i(\theta_i)=\mathbb{E}_{s,a,r,s'}\left[\left(y_i^{DQN}-Q(s,a;\theta_i)\right)^2\right]
\end{equation}
where
\begin{equation}
y^{DQN}_i=r+\gamma\max_{a'}Q(s',a';\theta^{-})
\end{equation}
$\theta^{-}$ denotes another identical architecture with delayed learning which is called \textit{target network}, an essential part of learning correct parameters in order to approximate Q function is to suspend target network's learning and update its parameters $\theta^{-}$ after a fixed number of iterations, while the other network $Q(s,a,\theta_i)$ is learning. Freezing parameters helps to eliminate moving target problem which leads to instability \cite{Mnih2015}. The gradient update for this network is:
\begin{dmath}
\nabla_{\theta_i}L_i(\theta_i)=\mathbb{E}_{s,a,r,s'}\left[\left(  y_i^{DQN}-Q(s,a;\theta_i)\right)\nabla_{\theta_i}Q(s,a;\theta_i) ) \right]
\end{dmath}
This algorithm does not require the model of the system considering the states and rewards are given by the environment, it is also off-policy since these states and rewards are not necessarily from the policy that is being trained. Another contributing factor the the success of DQN is \textit{experience replay} \cite{Lin1993,Mnih2015}. During the training process, we have a buffer which agents amass a data set of the episodes it's seen so far. Instead of just learning from current experience, agents randomly selects a mini-batch from this data set $\mathcal{D}_t$. This helps to increase the learning efficiency through re-iterating samples that have already been utilized and more importantly, eliminates the correlation amongst samples. 
\subsection{\small{The Dueling Network Architecture}}
\begin{figure}[!h]
\centering
\hbox{\hspace{1em} \includegraphics[scale=0.5]{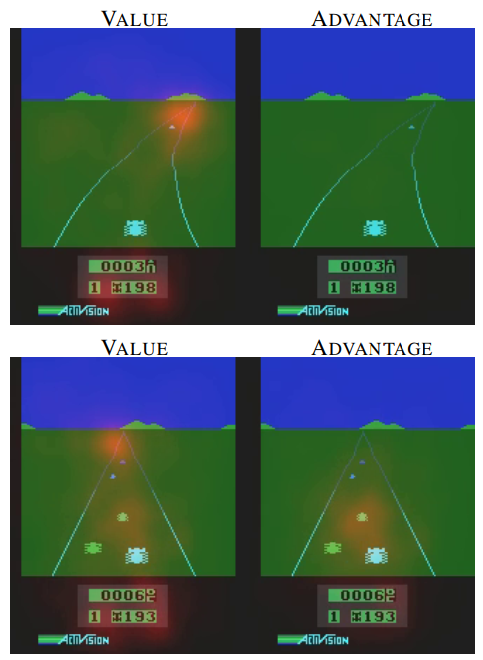}}
\caption{\footnotesize {Value and advantage saliency maps on Enduro. The value architecture learns to keep an eye out on the road while the advantage architecture gets high numeric values when a crash is imminent \cite{Wang2016}. }}
\label{fig3}
\end{figure}

\begin{figure}[!h]
\centering
\hbox{\hspace{0.3em} \includegraphics[scale=0.7]{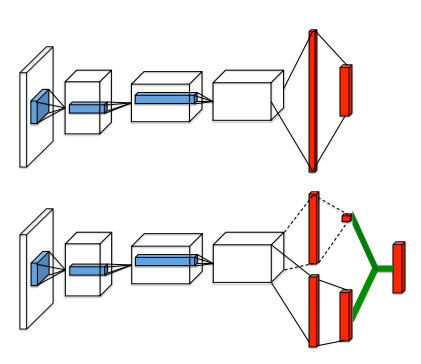}}
\caption{\footnotesize {Traditional one stream Q-network (top) and dueling network (bottom). This architecture separately approximates value and advantage functions, the green module implements Equation \ref{eq2} to blend the two together. Needless to say both networks output a set of values for Q function \cite{Wang2016}. }}
\label{fig4}
\end{figure}

Intuition behind dueling network architecture is that in many states, value of each action is not needed. as an example, in The Enduro game, moving to left or right is only important when an obstacle is on the road. In some states, it is of utmost importance to know which action to take, however, in other states it has no meaningful impact. This procedure is depicted in Figure \ref{fig3}. 
In order to implement this idea, a Q-network architecture was designed as illustrated in Figure \ref{fig4}. The first layers are convolutional layers like the original DQNs. having said that, instead of just a single stream of fully connected layers, there are two sequences of fully connected layers, which are designed in a way to approximate value and advantage functions separately. At the end, the two streams are incorporated together to estimate Q function, meaning like DQNs, the output is a set of values for Q function. Hence, it can be trained using well known algorithms such as SARSA. 

Blending these two streams requires a great deal of thought. we've already mentioned that $\mathbb{E}_{a\sim \pi(s)}[A^\pi(s,a)]=0$. Additionally, a deterministic policy, $a^*=\argmax_{a\in\mathcal{A}}Q(s,a)$ which entails $Q(s,a^*)=V(s)$, thus $A(s,a^*)=0$.

let's focus our attention on the architecture once more, one sequence of fully connected layers output a one-dimensional function $V(s;\theta,\beta)$, and the other a $|\mathcal{A}|$-dimensional array $A(s,a;\theta,\beta)$, which $\theta$ represents parameters of shared architecture (Convolutional layers) while $\alpha$ and $\beta$ are the parameters of the exclusive network for each (fully-connected layers).

One might be tempted to employ the definition of advantage and conclude:
\begin{equation}
Q(s,a;\theta,\alpha,\beta)=V(s;\theta,\beta)+A(s,a;\theta,\alpha)
\label{eq1}
\end{equation}
However, this equation applies to all state-action tuples, meaning we need to re-estimate $V(s,\theta,\beta), |\mathcal{A}|$ times. Moreover, Equation \ref{eq1} is unidentifiable in the sense that given Q we cannot approximate V and A in a unique way. This lack of identifiability is reflected by poor performance when this equation is used directly.

To address this problem, one can force the advantage function approximator to have zero advantage at the selected action. In other words, the last equation in the architecture handles the forward mapping
\begin{multline}
Q(s,a;\theta,\alpha,\beta)=V(s;\theta,\beta)+\\
\left(A(s,a;\theta,\alpha)-\max_{a'\in|\mathcal{A}|}A(s,a';\theta,\alpha)\right)
\end{multline}
 Now, for $a^*=\argmax_{a'\in\mathcal{A}}Q(s,a';\theta,\alpha,\beta)=\argmax_{a'\in\mathcal{A}}A(s,a';\theta,\alpha)$, we obtain $Q(s,a^*;\theta,\alpha,\beta)=V(s;\theta,\beta)$. Thus, the sequence $V(s;\theta,\beta)$ is an estimation for the value function, while the other is for advantage function. Alternatively, in order to have more stable training, max operator can be replaced with an average:
\begin{multline}
Q(s,a;\theta,\alpha,\beta)=V(s;\theta,\beta)+\\
\left(A(s,a;\theta,\alpha)-\frac{1}{|\mathcal{A}|}\sum_{a'}A(s,a';\theta,\alpha)\right)
\label{eq2}
\end{multline}
The original semantics of V and A is lost now that they're biased with a constant, but it's much easier to optimize, since the advantages change as fast as the mean, as opposed to compensating any changes to the optimal action's advantage in (12).
\section{Experiments} 
In this section, we go through mathematical aspects of the adjustment that we did to the dueling network architecture, which made an improvement to the original architecture.
\subsection{\small{Maximum Entropy}}
Maximum Entropy RL incorporates the entropy term with the reward, such that the optimal policy maximizes entropy for every state:
\begin{multline}
\pi^*_{MaxEnt}=\argmax_{\pi}\\
\sum_t\mathbb{E}_{s_t,a_t\sim \pi_s}[r(s_t,a_t)+\alpha\mathcal{H}(\pi(\cdot|s_t)]
\end{multline}
where $\alpha$ determines the relative significance of the entropy compared to the reward, which can be fine tuned based on the problem. This method unlike Boltzmann exploration \cite{Sallans2004} which greedily maximizes the entropy at current time, maximizes the entropy for the entire trajectory. Utilizing entropy has number of benefits, it offers superior exploration as it forces agent to visit states that have not been visited and leads to a more stable training \cite{Haarnoja2017}.

In our proposed method, we add the Entropy of Advantage function to our loss in order to improve exploration and achieve better results. 
\section{Results}
In this section, we discuss and shed light on the results which are illustrated in Table (1), it's essential to note that our proposed architecture was implemented on a single Nvidia GTX 1080 Ti graphics card and Intel Core i7-4770k CPU, leading to difference in scores compared to the Wang et al \cite{Wang2016}. 

As it's presented in Table (1), Maximum Entropy Dueling Network managed to increase the score by more than 10\% in both \textit{BankHeist} and \textit{RiverRaid}, Pong however was a different scenario, since it's a relatively simple game, with our current hardware, the score maxed out at 15 for all architectures. Forcing agent to explore different states by adding Entropy of Advantage function leads to a better exploration and ultimately, increased score.

\begin{center}
\captionof{table}{\footnotesize{Scores of Maximum Entropy Dueling Network in Pong, BankHeist and RiverRaid.}}
\begin{tabular}{c c c c} 
\toprule
Methods  & Pong & BankHeist & RiverRaid \\ [0.5ex] 
\midrule
ME & 15.1 & \textbf{385.3} & \textbf{3786.4} \\ 
DN & \textbf{15.2} & 345.2 & 3326.7\\ 
\bottomrule
\end{tabular}
\end{center}
\section{Conclusion}

In this paper, we introduced an improved architecture based on state-of-the-art Dueling Network Architecture for Atari 2600 domain, and shown that Maximum Entropy Dueling Network increases the score by more than 10\% and offers better exploration, having said that, we couldn't replicate the exact results presented in the original paper \cite{Wang2016} due to the lack of computational power. 
\section{References}

\bibliographystyle{IEEEtran}
\bibliography{refs}
\end{document}